\documentclass[preprint]{article}

\usepackage[nonatbib]{neurips_2024}

\usepackage[utf8]{inputenc} 
\usepackage[T1]{fontenc}    
\usepackage{hyperref}       
\usepackage{url}            
\usepackage{booktabs}       
\usepackage{amsfonts}       
\usepackage{nicefrac}       
\usepackage{microtype}      
\usepackage{xcolor}         
\usepackage{graphicx}
\usepackage{multirow}
\usepackage{booktabs}
\usepackage{tabularx}
\usepackage{amsmath}
\usepackage{amssymb}
\usepackage{array}
\usepackage{makecell}
\usepackage{comment}
\usepackage{colortbl}
\usepackage{amssymb} 
\usepackage{bbding} 
\usepackage{multirow} 
\usepackage{color} 

\usepackage{hhline}
\usepackage{boldline}
\usepackage{subfig}

\newcommand\eat[1]{}
\usepackage{array} 
\usepackage{enumerate} 
\usepackage{booktabs} 
\usepackage{framed} 
\usepackage{color} 
\usepackage{hyperref} 
\usepackage{booktabs}
\usepackage{array}
\usepackage{wrapfig}
\newcommand{\dataname}{\textsc{BabelBench}}
\newcommand{\nmodel}{13\;}
 
\usepackage{pifont} 

\title{\dataname: An Omni Benchmark for Code-Driven Analysis of Multimodal and Multistructured Data}

\author{
  \begin{tabular}[t]{c}
    Xuwu Wang, Qiwen Cui, Yunzhe Tao, Yiran Wang, Ziwei Chai, Xiaotian Han, Boyi Liu, \\
    Jianbo Yuan, Jing Su, Guoyin Wang, Tingkai Liu, Liyu Chen, Tianyi Liu, \\
    Tao Sun, Yufeng Zhang, Sirui Zheng, Quanzeng You, Yang Yang, Hongxia Yang \\
  \end{tabular} \\
  ByteDance Inc. \\
  \texttt{wangxuwu@bytedance.com}
}

\begin{document}

\maketitle

\begin{abstract}
  Large language models (LLMs) have become increasingly pivotal across various domains, especially in handling complex data types. This includes structured data processing, as exemplified by ChartQA and ChatGPT-Ada, and multimodal unstructured data processing as seen in Visual Question Answering (VQA). These areas have attracted significant attention from both industry and academia. Despite this, there remains a lack of unified evaluation methodologies for these diverse data handling scenarios. In response, we introduce \dataname, an innovative benchmark framework that evaluates the proficiency of LLMs in managing multimodal multistructured data with code execution. \dataname\ incorporates a dataset comprising 247 meticulously curated problems that challenge the models with tasks in perception, commonsense reasoning, logical reasoning, and so on. Besides the basic capabilities of multimodal understanding, structured data processing as well as code generation, these tasks demand advanced capabilities in exploration, planning, reasoning and debugging. Our experimental findings on \dataname\ indicate that even cutting-edge models like ChatGPT 4 exhibit substantial room for improvement. The insights derived from our comprehensive analysis offer valuable guidance for future research within the community. The benchmark data can be found at \url{https://github.com/FFD8FFE/babelbench}.
\end{abstract}

\section{Introduction}\label{sec:intro}
Large language models (LLMs) have demonstrated exceptional performance in a wide range of applications. 
For example, multimodal language models such as GPT-4V \cite{openai2023gpt4v}, Gemini \cite{geminiteam2024gemini}, and Claude \cite{anthropic2023claude3}; chart-specific models such as TableGPT \cite{li2023table} and ChartLlama \cite{han2023chartllama}; and code generation models represented by Codex \cite{chen2021evaluating}, Code Llama \cite{roziere2024code}, and DeepSeek-Coder \cite{2024deepseekcoder} have all demonstrated significant advancements. 


Building on these effectiveness in specific yet wide-ranging tasks, a genuinely intelligent LLM-as-Agent system should be capable of integrating these capabilities to address open-ended problems in the real world. Motivated by this vision, some systems and agents have been developed, such as BabyAGI \cite{babyagi2023}, LangChain \cite{Chase_LangChain_2022}, and AutoGPT \cite{yang2023autogpt}, which extend the application of LLMs to practical impacts in the physical world.

Despite these advancements, there is still a deficiency in benchmarks that comprehensively evaluate LLMs' performance in realistic and complex scenarios. 
Most existing benchmarks, such as SuperGLUE \cite{wang2020superglue},  MMLU \cite{hendrycks2021measuring} and MME \cite{fu2024mme}, mainly evaluate the acquisition of knowledge by LLMs and their ability to converse rather than solve problems. 
Although some work has introduced tasks that mimic real-world applications \cite{zhou2023webarena}, such as shopping on a website, the complexity of the involved data structure remains limited.

In this work, we propose that benchmarks should be based on multimodal and multistructured data. 
Beyond the inherent characteristics of real-world data, a unified understanding and processing of omni data is critical for addressing more complex tasks. For example, medical data comprises unstructured medical images, unstructured medical knowledge and dialogues, as well as structured treatment and laboratory data. Achieving a unified understanding and analysis of these diverse data types is essential for tasks such as lesion detection and disease prediction.
However, most works such as WebArena \cite{zhou2023webarena}, Infi-Agent \cite{hu2024infiagentdabench} and T-Eval \cite{chen2023t} are mainly limited to textual data. Towards multimodal applications, MiniWoB++ \cite{liu2018reinforcement}, WebShop \cite{NEURIPS2022_82ad13ec} and some other benchmarks \cite{drouin2024workarena,gao2024assistgui,koh2024visualwebarena} have integrated texts and images, but do not involve structured data. AgentBench \cite{liu2023agentbench} introduces both structured textual data such as database and semi-structured textual data such as knowledge graph, but all belong to the textual modality. Therefore, a multimodal and multistructured benchmark is critical.

\begin{figure}[t]
    \centering
    \includegraphics[width=0.8\linewidth]{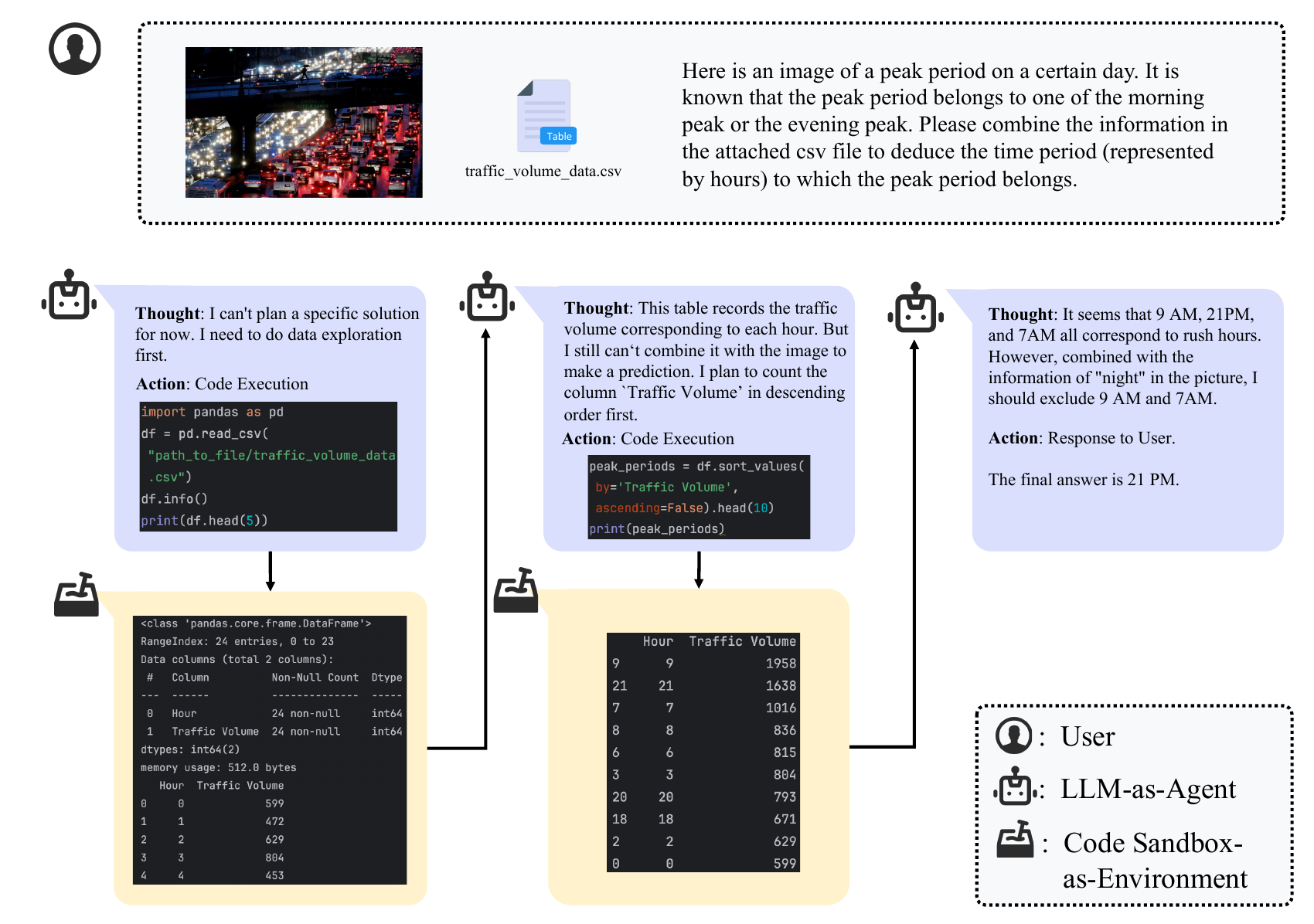}
    \caption{The problem prototype of \dataname.}
    \label{fig:workflow}
\end{figure}

Handling multimodal multistructured data faces many new challenges. Figure~\ref{fig:workflow} shows an example: the agent is given an image of traffic during rush hours, and a table containing the traffic volume at different times within a day. The agent is asked to determine the time period the image depicts.
Solving such a problem requires multiple advanced capabilities from the agent: 1) comprehending unified data structures and capturing the complicated alignment relationship. In this problem, the LLM needs to recognize that the "peak period" in the prompt corresponds to the columns "Hour" and "traffic volume" in the table. Additionally, the "Hour" values in the table should match the "night" attribute in the image. 2) Resolving complicated planning and reasoning. In this problem, the agent needs to decide what information (texts, images, structured data) is useful and in which order they should be processed. Compared to the simple data structure, the dependencies
between modalities are more complicated. In this problem, the agent needs to combine "the table shows that the candidates include 21 PM and 9 AM" and "the image excludes the answer of 9 AM" to deduce the final answer. 3) handling highly redundant information. Since images and tables usually contain more information, how to summarize and extract large amounts of information, as well as how to retain and understand large amounts of information is very challenging.


Therefore, to rigorously study the capabilities of LLM-as-Agent systems for tasks based on multi-modal and multistructured data, we propose a new benchmark \dataname. It consists of 247 human-annotated questions based on textual questions, images and structured tables. From the perspective of LLMs' capabilities, it assesses the capabilities of multimodal understanding, table interpretation and code generation. From a task completion perspective, it requires perceptual skills, including counting, color recognition, and optical character recognition (OCR), alongside reasoning abilities such as commonsense reasoning, spatial reasoning, and logical reasoning. Regarding the dataset quality, all annotations are performed by domain experts, involving features such as debiasing traps, deep domain knowledge, and multi-step complex reasoning.  As for evaluation confidence, the benchmark includes three levels of difficulty—easy, medium, and hard—to ensure a balanced evaluation across diverse capability dimensions, supplemented by a robust automated evaluation framework. 

Overall, our contributions can be summarized as follows:
\begin{itemize}
    \item To the best of our knowledge, this is the first work to address the importance of multimodal and multistructured data in terms of LLM evaluation. In addition, the proposed benchmark \dataname\  satisfies high quality, good diversity, and appropriate difficulty.
    \item To support existing models, we have implemented a complete evaluation framework, including components such as file/image processing, code execution, data analysis, and automated effect verification. The dataset and framework will be released at github under the license of CC-BY-4.0 \footnote{It will also be provided in the supplemental materials.}.
    \item Through experiments with \nmodel LLMs, we discover that even ChatGPT 4 has substantial room for improvement on \dataname. The insights derived from our comprehensive analysis offer valuable guidance for future research within the community.
\end{itemize}

\section{Design and Development}\label{sec:bench}


\begin{figure}[t]
    \centering
    \includegraphics[width=1.0\linewidth]{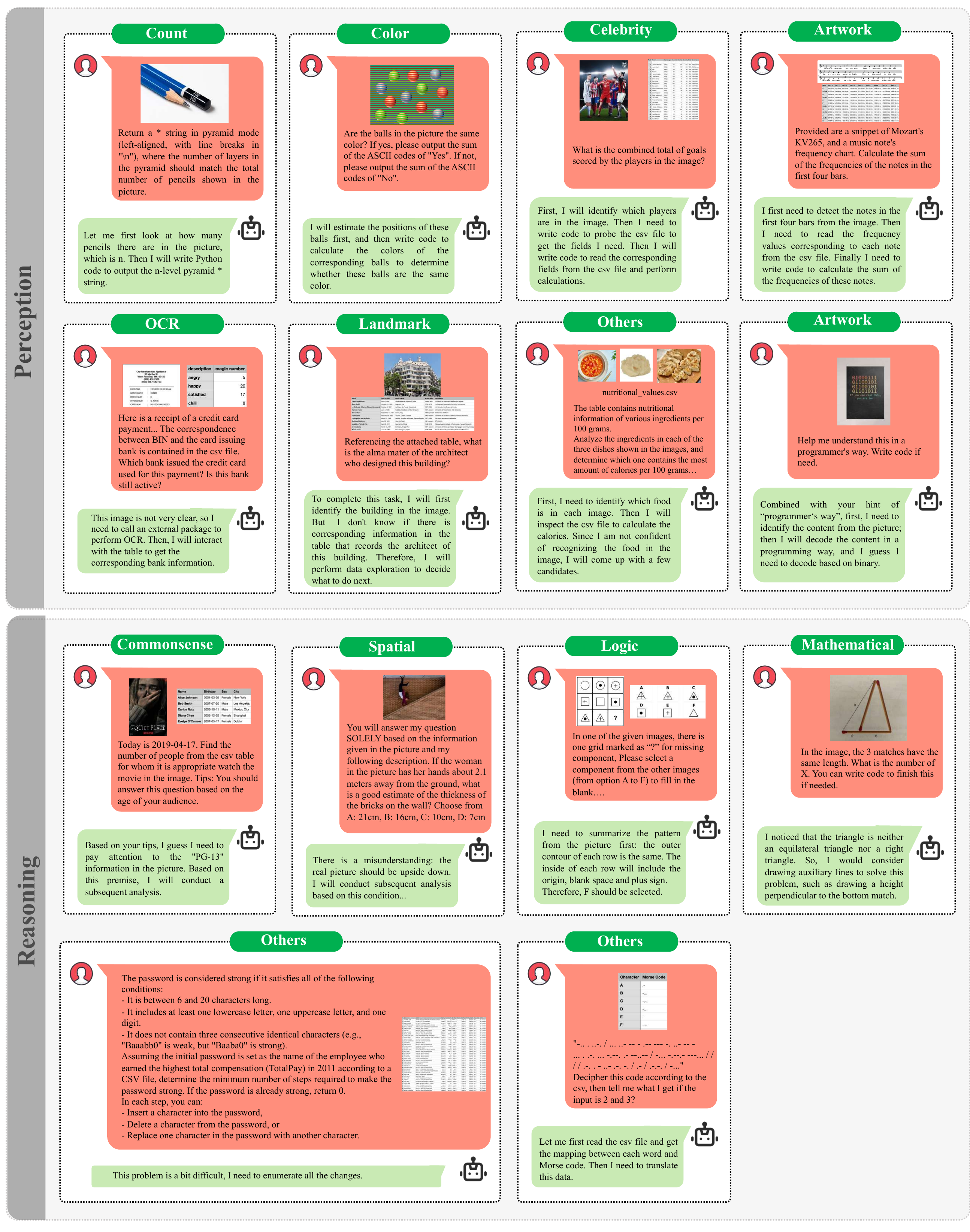}
    \caption{Examples of \dataname.}
    \label{fig:eg_of_benchmark}
\end{figure}

\subsection{Design}\label{sec:design}

Real-world problems are inherently open-ended, which are not suitable for systematic evaluation. To address it, we abstract real-world scenarios from two critical perspectives: the model-inherent capabilities and the task-dependent capabilities. This abstraction allows us to encapsulate real-world problems within a smaller but more manageable scope.


\noindent \textbf{Model-inherent capabilities} As a prerequisite, we identify two essential model capabilities from real-world tasks. First, to align with typical data types in the real world, LLMs should be able to comprehensively understand, analyze and process the unified multimodal multistructured data. In addition, to enhance LLMs' impact on the physical world, LLMs should interact with an environment through tools, involving both structured data and images. Specifically, in this paper, we represent tool-using with the generation and execution of code. We require the presence of a sandbox environment to execute code, process text and images, and return execution results (including outputs and errors).
Therefore, an LLM/agent needs to have basic capabilities in multimodal understanding, tabular data processing, and programming in order to obtain an entry ticket to \dataname.

\noindent \textbf{Task-dependent capabilities} To ensure the diversity and comprehensiveness of our dataset, inspired by \cite{fu2024mme}, we then establish a detailed capability taxonomy to guide the annotation. As presented in Figure~\ref{fig:eg_of_benchmark}, it mainly consists of perception and reasoning, each encompassing multiple subcategories. Besides, to increase the difficulty, we incorporate a variety of complex factors and challenges for each subcategory, as listed in Table~\ref{tbl:difficulty}.
\begin{table}[h]
\centering
\caption{The designed difficulties for each subcategory.}
\small
\label{tbl:difficulty}
\begin{tabular}{p{1.5cm}|p{11cm}}
\toprule
\multicolumn{2}{l}{\cellcolor{gray!30}\textbf{Perception}}\\ \hline
\midrule
\textbf{Count} & counting of objects with occlusions (cf. Figure~\ref{fig:eg_of_benchmark}), large quantities of objects, etc. \\ \hline 
\textbf{Color} & recognition of misleading colors (cf. Figure~\ref{fig:eg_of_benchmark}), colors alongside other attributes, etc. \\ \hline 
\textbf{Celebrity} & recognizing multiple objects (cf. Figure~\ref{fig:eg_of_benchmark}), mapping logos to corresponding entities, etc. \\ \hline 
\textbf{Artwork} & analysis on artworks such as paintings, sculpture and music (cf. Figure~\ref{fig:eg_of_benchmark}). \\ \hline 
\textbf{OCR} & recognizing confusing characters (cf. Figure~\ref{fig:eg_of_benchmark}), OCR with post-processing, etc. \\ \hline 
\textbf{Landmark} & analysis of uncommon landmarks (cf. Figure~\ref{fig:eg_of_benchmark}) and expanding related knowledge. \\ \hline 
 \midrule

\multicolumn{2}{l}{\cellcolor{gray!30}\textbf{Reasoning}}\\ \hline \midrule
\textbf{Common} & reasoning about social norms (cf. Figure~\ref{fig:eg_of_benchmark}), physical common sense, game rules, etc. \\ \hline 
\textbf{Spatial} & reasoning about rare spatial view (cf. Figure~\ref{fig:eg_of_benchmark}), relative spatial positions, etc. \\ \hline 
\textbf{Logic} & reasoning about induction (cf. Figure~\ref{fig:eg_of_benchmark}), deduction, analogy, etc.\\ \hline 
\textbf{Math} & reasoning about geometry (cf. Figure~\ref{fig:eg_of_benchmark}), algebra, and their integration with coding. \\
\bottomrule
\end{tabular}
\end{table}

\subsection{Construction}\label{sec:construction}

The construction process consists of two steps, outlined as follows:

First, 15 experts were engaged for annotation. Comprehensive annotations were required for each question, encompassing the question itself, the data sources, the ground-truth answer, and the evaluation criteria. To guarantee the annotation consistency, each question then underwent a quality review by a second expert. After this round of annotation and review, we obtained 491 questions in total.

After that, 5 experts performed a secondary review to choose 10 to 30 questions from each subcategory, forming the ultimate benchmark. This selection was influenced by factors such as question difficulty, the range of model capabilities required, the potential for data leakage, etc. Finally, 247 questions formed the final dataset.


\subsection{Evaluation Framework}\label{sec:evaluation}

To support an end-to-end evaluation of existing LLMs, we also provide an evaluation pipeline. In alignment with React \cite{yao2023react}, it can integrate either locally deployed models or remote open model APIs. Besides, it proficiently supports the reading and manipulation of images and tables, while also enabling environmental interaction through the tool of Python sandbox. Specially, since some model do not support large images and multiple images, we also support to stitch multiple images together and down-sample the images.

To ensure consistent evaluation results over time, we also propose an automated evaluation metric calculation solution. By specifying the variable names and data types as constraints in the question, we can extract the prediction from LLM's response and evaluate it based on the specified data types and matching methods. Additionally, we have extended the evaluation to support a wider range of answer data types and matching methods, accommodating scenarios such as multiple correct answers, answers requiring expression execution, numerical matching with precision requirements, etc. The supported evaluation solution is listed in Figure~\ref{fig:enter-label} and Table~\ref{table:enter-label}. To implement it, annotators were asked to systematically classify the answers' data type and evaluation method. 
Finally, we report the overall accuracy of each question.

\begin{minipage}{0.4\textwidth}
    \centering
    \includegraphics[width=0.9\linewidth]{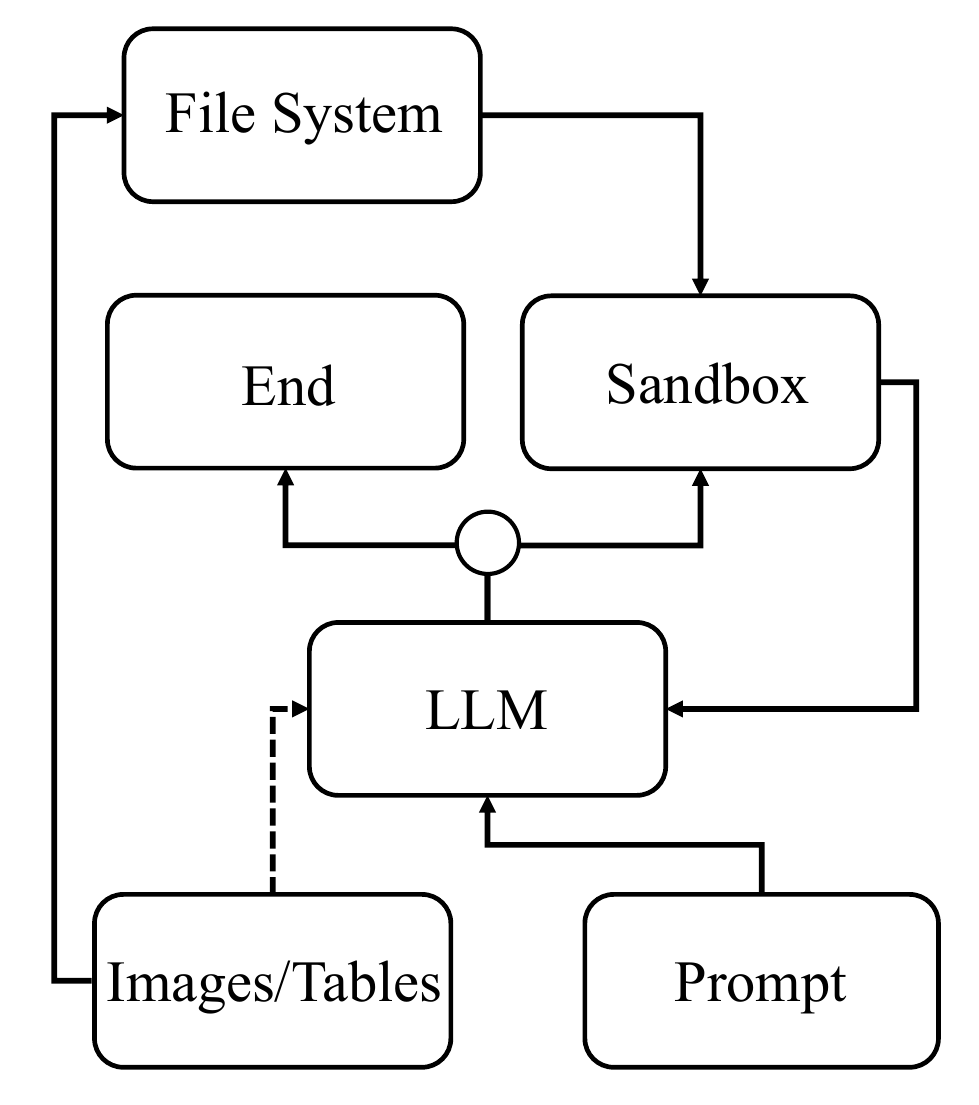}
    \captionof{figure}{The prototype of the evaluation framework. The data flow corresponding to the dashed line is optional, depending on whether the model supports it.}
    \label{fig:enter-label}
\end{minipage}
\hspace{0.25cm}
\begin{minipage}{0.55\textwidth}
\centering
\captionof{table}{The supported variable types and matching methods during evaluation.}
\label{table:enter-label}
\setlength{\tabcolsep}{5pt} 
\footnotesize
\begin{tabular}{p{1.2cm}p{1.3cm}p{4.5cm}}
\toprule
\textbf{Variable} & \textbf{Matching} & \textbf{Description} \\
\midrule
\multirow{4}{*}{\textbf{str}} & exact & Exact matching. \\
& fuzzy & Ignore case and match based on `str.contains()'. \\
& either\_ok & Either of correct answers is ok. \\
& execution & Execute the expression and exactly match the execution results. \\
\midrule
\multirow{1}{*}{\textbf{float}} & - & Floating point matching based on tolerance. \\
\midrule
\textbf{bool} & - & Exact matching. \\
\midrule
\textbf{int} & - & Exact matching. \\
\midrule
\textbf{list}, \textbf{list\_of\_int}, \textbf{list\_of\_str} & exact & Exact matching including the order. \\
& disordered & Order of elements is not important. \\
\midrule
\textbf{list\_of\_list} & - & Exact matching. \\
\bottomrule
\end{tabular}
\end{minipage}

\section{Experiments}\label{sec:exp}
\subsection{Experimental Setup}

\paragraph{Multimodal LLMs as agents} \nmodel models/systems are included for evaluation. They can be roughly classified into:
\begin{itemize}
    \item \textbf{Commercial System (System)}: So far, only OpenAI's ChatGPT can independently evaluate this benchmark. Other systems (Claude, Gemini, Qwen, etc.) lack full support for code execution, multiple image uploads, and multiple CSV file uploads. Therefore, we only include ChatGPT 4 in this category, with data manually collected from 2024/05/11 to 2024/05/15.
    \item \textbf{Closed-source LLMs (CSS)}: We select some widely recognized models along with their different size versions: GPT-4 \cite{gpt4}, GPT-4o \cite{gpto}, QWen-VL-Plus \cite{bai2023qwenvl}, QWen-VL-Max, Gemini-pro-1.0 \cite{geminiteam2024gemini}, Gemini-pro-1.5, Claude3-Sonnet \cite{anthropic2023claude3}, Claude3-Haiku, Claude3-Opus.
    \item \textbf{Open-source LLMs (OSS)}: We also selected several open-source models for evaluation. However, due to some models' limited ability to follow complex instructions, our evaluation framework could not effectively assess them. Ultimately, the models we report include InternXComposer2 \cite{dong2024internlmxcomposer2}, InternVL \cite{chen2023internvl}, and LLaVa \cite{liu2024improved}.
\end{itemize}

\paragraph{Implementation details} We implement the entire framework based on the work of Infi-Agent~\cite{hu2024infiagentdabench} following the license of Apache 2.0. For CSS LLMs and OSS LLMs, the experiments are conducted based on our framework. First, we obtain the API for the corresponding CSS LLMs or deploy the OSS LLMs. Next, we integrate them into our pipeline. To reduce the number of additional interaction rounds, if the problem involves a table, we provide a schema display of that table in the prompt. Other details can be found in Table~\ref{tbl:backbone_info}.
\begin{table}[t]
\centering
\caption{Evaluation details of the \nmodel LLMs.} \label{tbl:backbone_info}
\footnotesize
\begin{tabular}{p{1cm}p{3cm}p{0.8cm}p{1.6cm}p{2.4cm}p{2cm}}
\toprule
\textbf{Cate.} & \textbf{Model} & \textbf{Size} & \textbf{Version} & \textbf{Creator} & \textbf{Collection} \\
\midrule
System & ChatGPT 4 & N/A & 2024/05/11-2024/05/15 & OpenAI & Manually \\ 
\midrule
\multirow{10}{*}{CSS} & GPT-4 \cite{gpt4} & N/A & 20230613 & OpenAI & API on Azure\\
& GPT-4o \cite{gpto} & N/A & 20240513 & OpenAI & API on Azure \\
& QWen-VL-Plus \cite{bai2023qwenvl} & N/A & - & Alibaba & API on Aliyun \\
& QWen-VL-Max  \cite{bai2023qwenvl} & N/A & - & Alibaba & API on Aliyun \\
& Gemini-pro-1.0 \cite{geminiteam2024gemini} & N/A & v1  & Google & Official API\\
& Gemini-pro-1.5 \cite{geminiteam2024gemini} & N/A & v1  & Google & Official API \\
& Claude3-Sonnet \cite{anthropic2023claude3}&  N/A & 20240229  & Anthropic & Official API \\
& Claude3-Haiku \cite{anthropic2023claude3}& N/A & 20240307 & Anthropic & Official API \\
& Claude3-Opus  \cite{anthropic2023claude3}& N/A & 20240229 & Anthropic & Official API \\
\midrule
\multirow{3}{*}{OSS} & LLaVa \cite{liu2024improved} & 34B & V1.6 & UW-Madison et al. & Self-hosting \\
& InternXComposer2 \cite{dong2024internlmxcomposer2} & 7B & V2 & OpenGVLab & Self-hosting  \\
& InternVL \cite{chen2023internvl} & 25.5B & V1.5 & OpenGVLab & Self-hosting \\
\bottomrule
\end{tabular}
\end{table}

\subsection{Main Results}

\begin{table}[thb]
\centering
\caption{Performance on \dataname. The best results are \textbf{bolded}, and the second-best results are \underline{underlined}. For "Has file?", "I\&T" indicates the presence of both images and tables, while "Has table?-No" and "Has img?-No" correspond to cases with only images and only tables, respectively.} \label{tbl:performance}
\setlength{\tabcolsep}{6pt} 
\begin{tabular}{p{0.6cm}p{2.3cm}p{0.6cm}p{0.6cm}p{0.6cm}p{0.6cm}p{0.6cm}p{0.6cm}p{0.6cm}p{0.6cm}>{\centering\arraybackslash}p{0.6cm}p{0.6cm}p{0.6cm}p{0.6cm}}
\toprule
\multirow{2}{*}{\textbf{Type}} & \multirow{2}{*}{\textbf{LLM}} & \multirow{2}{*}{\textbf{Acc}} & \multicolumn{2}{c}{\textbf{Has image?}} & \multicolumn{2}{c}{\textbf{Has table?}} & 
\multicolumn{1}{c}{\textbf{Has file?}} & 
\multicolumn{3}{c}{\textbf{Difficulty}} \\ 
\cline{4-11}
& & & \textbf{Yes} & \textbf{No} & \textbf{Yes} & \textbf{No} & \multicolumn{1}{c}{\textbf{I\&T}} & \textbf{Easy} & \textbf{OK} & \textbf{Hard} \\
\midrule
System & ChatGPT 4 & \textbf{42.11} & \textbf{41.53} & \textbf{54.55} & \textbf{40.91} & \textbf{43.48} & \multicolumn{1}{c}{\textbf{39.67}} & \textbf{55.93} & \textbf{39.81}& \textbf{35.29} \\
\midrule
\multirow{10}{*}{CSS} & GPT-4 & 31.98 & 30.93 & \textbf{54.55} & \underline{38.64} & 24.35 & \multicolumn{1}{c}{37.19} &  \underline{49.15} & 28.16 & 24.71 \\
& GPT-4o & 32.39 & 32.63 & 27.27 & 31.06 & \underline{33.91} & \multicolumn{1}{c}{31.40} & 47.46 & 29.13 & 25.88 \\
& {\small QWenVL-plus} & 1.21 & 1.27 & 0.00 & 0.00 & 2.61 & \multicolumn{1}{c}{0.00} &  1.69 & 0.97 & 1.18 \\
& {\small QWenVL-max} & 12.15 & 12.29 & 9.09 & 7.58 & 17.39 & \multicolumn{1}{c}{7.44} &  13.56 & 10.68 & 12.94 \\
& {Gemini-pro-1.0} & 17.0 & 16.53 & 27.27 & 15.91 & 18.26 & \multicolumn{1}{c}{14.88} & 28.81 & 15.53 & 10.59 \\
& {Gemini-pro-1.5} & \underline{34.01} & \underline{33.90} & \underline{36.36} & 37.88 & 29.57 & \multicolumn{1}{c}{\underline{38.02}} & 47.46 & \underline{32.04} & \multicolumn{1}{c}{\underline{27.06}} \\
& {Claude3-sonnet} & 7.29 & 7.20 & 9.09 & 5.30 & 9.57 & \multicolumn{1}{c}{4.96} & 15.25 & 5.83 & 3.53 \\
& {Claude3-haiku} & 6.48 & 6.36 & 9.09 & 3.79 & 9.57 & \multicolumn{1}{c}{3.31} & 5.08 & 8.74 & 4.71 \\
 & {Claude3-opus} & 22.67 & 22.46 & 27.27 & 21.97 & 23.48 & \multicolumn{1}{c}{21.49} & 32.20 & 27.18 & 10.59 \\
\midrule
\multirow{3}{*}{OSS} & {InternX} & 4.45 & 4.66 & 0.00 & 2.27 & 6.96 & \multicolumn{1}{c}{2.48} & 1.69 & 4.85 & 5.88 \\
& {InternVL} & 7.69 & 7.20 & 18.18 & 5.30 & 10.43 & \multicolumn{1}{c}{4.13} & 11.86 & 6.80 & 5.88 \\
& {LLaVa} & 4.45 & 3.81 & 18.18 & 5.30 & 3.48 & \multicolumn{1}{c}{4.13} & 11.86 & 1.94 & 2.35 \\
\bottomrule
\end{tabular}
\end{table}

Table~\ref{tbl:performance} presents the key experimental results, revealing several critical insights. 

\noindent \textbf{Overall performance} ChatGPT 4 achieves the highest score, but a score of 42.11 highlights substantial room for improvement. Among closed-source models, Gemini-Pro-1.5 achieves the highest score, followed by GPT-4o and GPT-4. Despite these performances, a noticeable gap remains compared to ChatGPT 4, indicating that additional data matching the execution environment could be beneficial. Additionally, there is a significant disparity between closed-source and open-source models, suggesting that the open-source community still has much progress to make.

\noindent \textbf{Influence of multistructured data} Regarding the impact of images, ChatGPT 4, GPT-4, and Gemini perform better on questions without images, while models like QwenVL-max and GPT-4o perform better on questions with images. 
Considering the impact of tables, ChatGPT 4, GPT-4o, and Claude3-opus exhibit balanced performance, whereas GPT-4 and Gemini-1.5 perform significantly better on data with tables. 
This indicates that current models do not optimize for modality and data structure uniformly. 
For the combined influence of images and tables, we initially predicted that questions with both would be the most challenging and thus yield the lowest performance. 
However, we found that models like GPT-4o and Gemini-Pro-1.5 did not conform to this hypothesis. 
We speculate that this may be due to the information gain brought by different modalities and data structures.



\begin{figure}[h!]
    \centering
    \begin{minipage}[b]{0.4\textwidth}
        \centering
        \includegraphics[width=\textwidth]{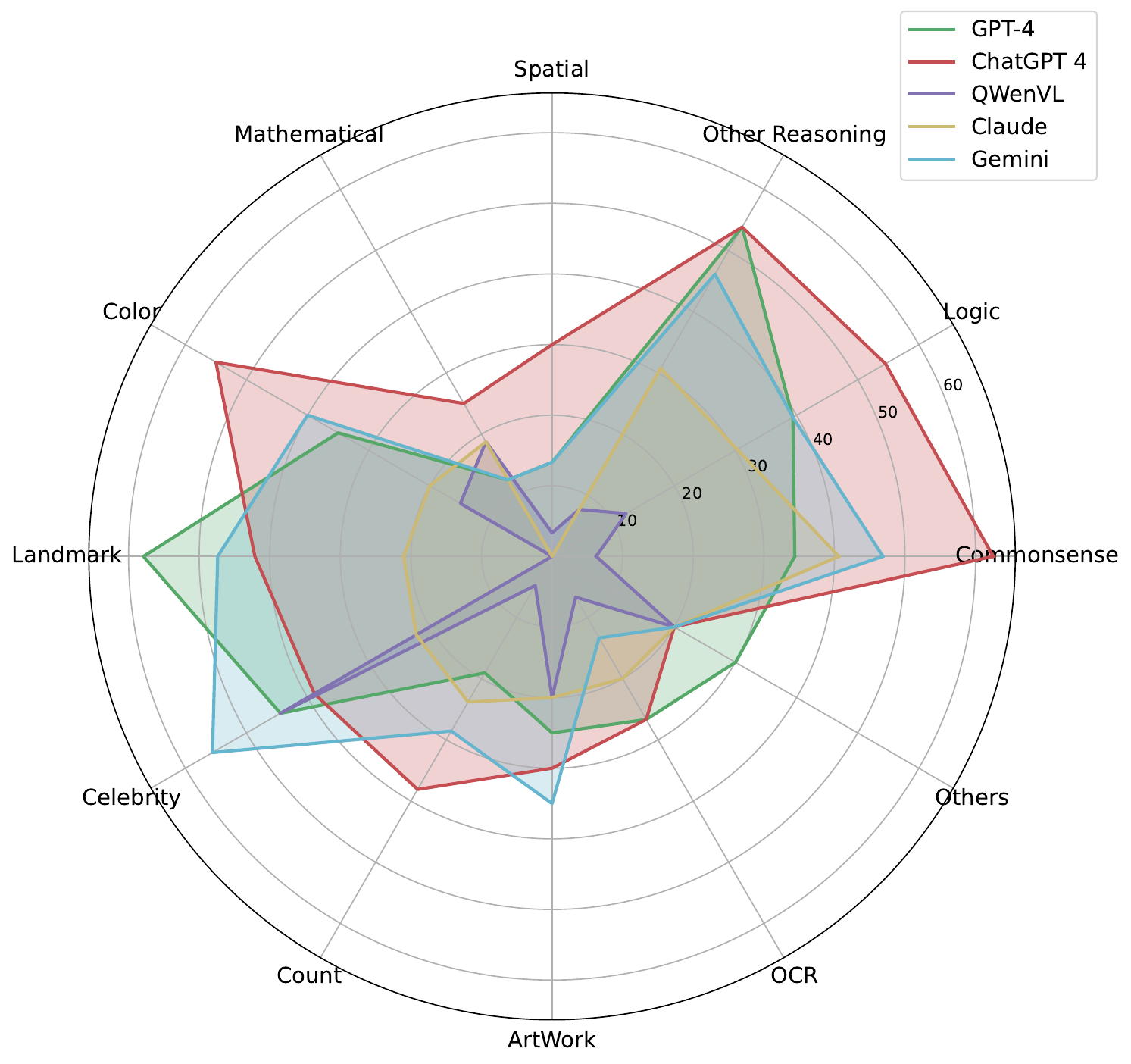}
        \caption{Model performance in terms of capabilities.}
        \label{fig:perf_ability}
    \end{minipage}
    \hfill
    \begin{minipage}[b]{0.45\textwidth}
        \centering
        \includegraphics[width=\textwidth]{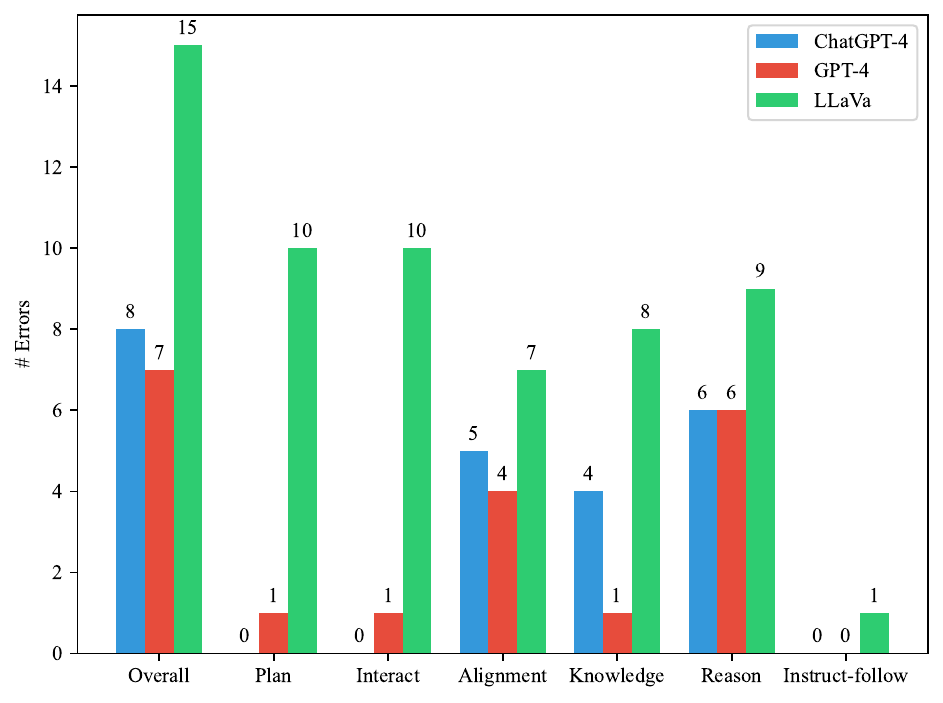}
        \caption{Error analysis of 3 representative LLMs.}
        \label{fig:error_analysis}
    \end{minipage}
\end{figure}

\noindent \textbf{Task difficulty} 
As expected, most models score higher on easy tasks than on medium tasks, and higher on medium tasks than on hard tasks. However, the highest score for easy tasks was only 55.93, revealing significant deficiencies even in handling simple tasks.

\noindent \textbf{Capability} We also select a representative subset of models to compare their performance in terms of different capabilities. According to Figure~\ref{fig:perf_ability}, all LLMs perform poorly on Spatial Reasoning, Mathematical Reasoning, Counting, Artwork and ORC. This suggests the direction for future optimizations.

\subsection{Detailed Analysis}

To delve deeper into the effectiveness of LLMs on \dataname, we undertake an extensive case study. Specifically, we establish several key aspects for evaluation:
1) \textbf{Overall}: Whether the solution effectively addresses the given question. 2) \textbf{Plan}: Assessing the model’s ability to decompose the question intent and reasonably arrange the dependencies among the sub-steps. 3) \textbf{Interact}: Analyzing the model’s interaction with the external environment, including active engagement to extract necessary information from images and tables, refinement of the plan based on interaction results if the initial plan is flawed or absent, and debugging and fixing corner cases through external feedback. 4) \textbf{Alignment}: Evaluating the ability to align and interact with information across prompts, images, and tables. 
5) \textbf{Knowledge}: Assessing the model's possession of essential common sense and domain-specific knowledge. 6) \textbf{Reasoning}: Testing the model's reasoning capabilities, such as inducing rules and detecting outliers that contradict common patterns. 
7) \textbf{Instruct-follow}: Determining whether the model adheres to formatting guidelines specified in the prompt. 

Based on this standard, we randomly select 15 questions and utilize ChatGPT 4, GPT-4, and LLaVa as representative models for case study. Specifically, for each aspect, scores of 0, 0.5, 1, or 'unknown' are assigned. It should be noted that due to preliminary errors, some aspects may cannot be evaluated in the answers\footnote{This means that if there are few errors in a aspect, it does not necessarily indicate good performance in this aspect, but rather that it may be difficult to assess errors in it.}. The number of errors each model made is illustrated in Figure~\ref{fig:error_analysis}.

From the statistical results and the cases' details, we find out that:
1) For both ChatGPT 4 and GPT-4, errors predominantly occur in the alignment, knowledge and reasoning aspects, while LLaVa displays profound deficiencies across all tested abilities. 2) For alignment, the errors primarily concentrate on: a) mapping text objects with complex constraints to objects in images; b) aligning values or analytical conclusions of tables with images. 3) For knowledge, commonsense knowledge has been well mastered, but domain-specific knowledge is still insufficient. 4) In terms of reasoning, they still lack in mathematical reasoning, pattern induction, and trap avoidance.


\subsection{Other Findings}

When data becomes multimodal and multistructured, we discover that the LLMs encounter many new findings.

\begin{figure}
    \centering
    \includegraphics[width=0.8\linewidth]{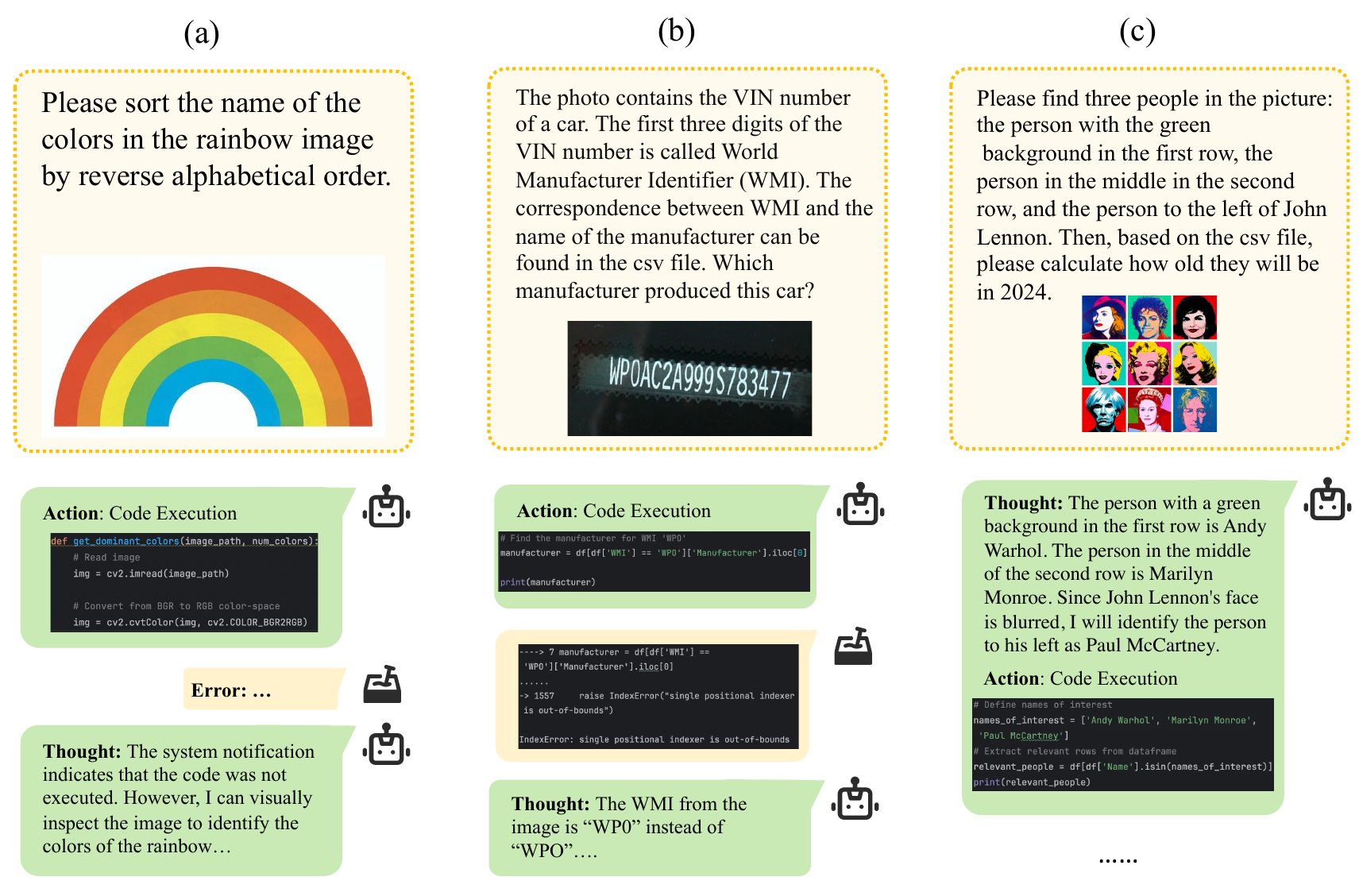}
    \caption{Examples to support the insights. The responses are collected with GPT-4.}
    \label{fig:eg_of_tool}
\end{figure}

\noindent\textbf{The self-efficacy of LLMs needs further improvement.} For each problem or its subcomponents, the model can either rely on its inherent capabilities (such as image/table inspection) or leverage tools (such as code). When handling complex tasks, it is crucial for the model to allocate cognitive load based on task requirements and tool availability, deciding whether to rely on itself or on external tools. However, we have found that current LLMs are not yet capable of effectively making this decision. For instance, in Figure~\ref{fig:eg_of_tool} (a), for the task of "identifying colors from the image", the model initially attempts to use code, only switching to direct image inspection upon encountering errors. However, direct image observation is sufficient to solve the problem with a low error probability, whereas using code is more complex and error-prone.

\noindent\textbf{LLMs have a low adversity quotient.} 
An error triggered by one data type may actually be caused by another data type. Therefore, the model should be aware that errors may originate from earlier steps and data sources; and it can also leverage other data sources to resolve the errors.
For example, in Figure~\ref{fig:eg_of_tool} (b) requiring OCR from the image and corresponding data retrieval from the table, the LLM misinterprets "WP0" in the image as "WPO", leading to an index error when searching the table. To deal with it, GPT-4 made no corrections in its first two attempts but silently corrected the error on the third try. However, a more reasonable solution is to inspect the table content to recognize the presence of "WP0", and thereby correct the error in the OCR.

\noindent\textbf{Starting from the right data source may simplify the task}. Anchoring on different data sources can result in varying levels of task complexity. For instance, consider Figure~\ref{fig:eg_of_tool} (c): if LLMs start by anchoring on images, identifying the individuals in images is relatively challenging. However, if they begin by anchoring on a table, the task becomes easier. The table provides a list of individuals, which drastically narrows down the pool of candidates, thereby reducing the problem's difficulty.

\section{Related Work}\label{sec:related_work}
\noindent \textbf{Benchmarking multimodal agents} The rapid advancement of multimodal large language models (MLLMs) has led to an increased emphasis on benchmarks for evaluating their performance across various domains. Existing benchmarks, such as MMBench~\cite{liu2024mmbench}, SEED-Bench~\cite{li2023seedbench}, SEED-Bench-2~\cite{li2023seedbench2}, TableVQA~\cite{kim2024tablevqabench} and MathVista~\cite{lu2024mathvista}, assess MLLMs across a range of ability dimensions, including image-text understanding, scene understanding, detection, optical character recognition (OCR), and mathematical reasoning, but lack a comprehensive investigation specifically tailored for an omni-benchmark that includes code-driven analysis of structured data like tables and unstructured data like images. In the realm of digital interaction agents, benchmarks span coding environments, web scenarios~\cite{zhou2024webarena,yao2023webshop,deng2023mind2web,chen2021websrc} and mobile applications~\cite{burns2021mobile,sun2022metagui,rawles2023android,humphreys2022datadriven,li2020mapping}. But these efforts often focus on singular environments or lack executability. 

\noindent \textbf{Benchmarking code-driven tasks}
Recent years have seen the emergence of LLMs that exhibit impressive code generation and analysis capabilities. Various benchmarks have been proposed to assess LLM performance on code-driven tasks. But evaluating these models comprehensively remains an open challenge. CodeXGLUE~\cite{lu2021codexglue} is a benchmark covering both code understanding and generation tasks. Other benchmarks, such as those by~\cite{chen2021evaluating,austin2021program,huang2023reasoning,Li_2022}, focus on code generation in competition-level settings. DS-1000~\cite{lai2022ds1000} and the dataset by~\cite{bai2023qwen} assess LLMs' ability in data science and general-purpose tasks. Research on practical LLM-based code completion tools, such as Github's Copilot~\cite{github22}, has characterized interaction models~\cite{xu2021inide,barke2022grounded} and user experiences~\cite{ziegler2022productivity}, highlighting both benefits and potential drawbacks~\cite{ziegler2022productivity}. Despite progress in LLM benchmarks for code-driven tasks, a comprehensive benchmark covering structured and unstructured multimodal data analysis is still needed. 


\section{Limitations}
Our dataset has certain limitations: Firstly, due to the high difficulty level of each task and the high annotation cost, its scale is relatively limited. Future work could expand the dataset to include more samples. Secondly, for systematic and automated evaluation, there is a certain gap between our tasks and real-world scenarios. Future work could consider designing tasks that are more reflective of real-world situations.

\section{Conclusion}\label{sec:conclusion}
In this paper, we present a new benchmark \dataname\ designed to evaluate large language models in multimodal and multistructured data scenarios. By incorporating 247 human-annotated questions, \dataname\ assesses LLM capabilities in multimodal understanding, table interpretation, and code generation, while also testing perceptual and reasoning skills. Our experiments with \nmodel LLMs, including ChatGPT 4, reveal substantial room for improvement, underscoring the complexity of the tasks. \dataname\ sets a new standard for comprehensive LLM evaluation, guiding future research towards the development of more intelligent and versatile LLM-as-Agent systems capable of addressing real-world challenges.

\clearpage
{\small\bibliography{neurips_2024}}

\clearpage

\end{document}